\documentclass[conference]{IEEEtran}
\IEEEoverridecommandlockouts
\usepackage{cite}
\usepackage{amsmath,amssymb,amsfonts}
\usepackage{algorithmic}
\usepackage{graphicx}
\usepackage{textcomp}
\usepackage{xcolor}
\usepackage{booktabs} % For formal tables
\usepackage{graphicx}
\usepackage{subcaption}
\usepackage{multirow}
\usepackage{algorithm2e}
\usepackage{tabularx}
\usepackage{tabulary}
\usepackage{cleveref}
\usepackage{float}
\usepackage{amsmath}
\usepackage{makecell}
\usepackage{color}
\def\BibTeX{{\rm B\kern-.05em{\sc i\kern-.025em b}\kern-.08em
    T\kern-.1667em\lower.7ex\hbox{E}\kern-.125emX}}
\begin{document}

\title{GISNet:Graph-Based Information Sharing Network For Vehicle Trajectory Prediction}

\author{\IEEEauthorblockN{Ziyi Zhao, Haowen Fang, Zhao Jin, Qinru Qiu}
\IEEEauthorblockA{Department of Engineering and Computer Science, Syracuse University, Syracuse, New York, USA\\
Email: {\{zzhao37, hfang02, zjin04, qiqiu\}@syr.edu} }
}
\maketitle
\begin{abstract}
The trajectory prediction is a critical and challenging problem in the design of an autonomous driving system. Many AI-oriented companies, such as Google Waymo, Uber and DiDi, are investigating more accurate vehicle trajectory prediction algorithms. However, the prediction performance is governed by lots of entangled factors, such as the stochastic behaviors of surrounding vehicles, historical information of self-trajectory, and relative positions of neighbors, etc. In this paper, we propose a novel graph-based information sharing network (GISNet) that allows the information sharing between the target vehicle and its surrounding vehicles. Meanwhile, the model encodes the historical trajectory information of all the vehicles in the scene. Experiments are carried out on the public NGSIM US-101 and I-80 Dataset and the prediction performance is measured by the Root Mean Square Error (RMSE). The quantitative and qualitative experimental results show that our model significantly improves the trajectory prediction accuracy, by up to 50.00\%, compared to existing models.

\end{abstract}


% \fi
\IEEEpeerreviewmaketitle

\section{Introduction}\label{sec:introduction}

\begin{figure}[t]
\centering
\includegraphics[width=3.4in]{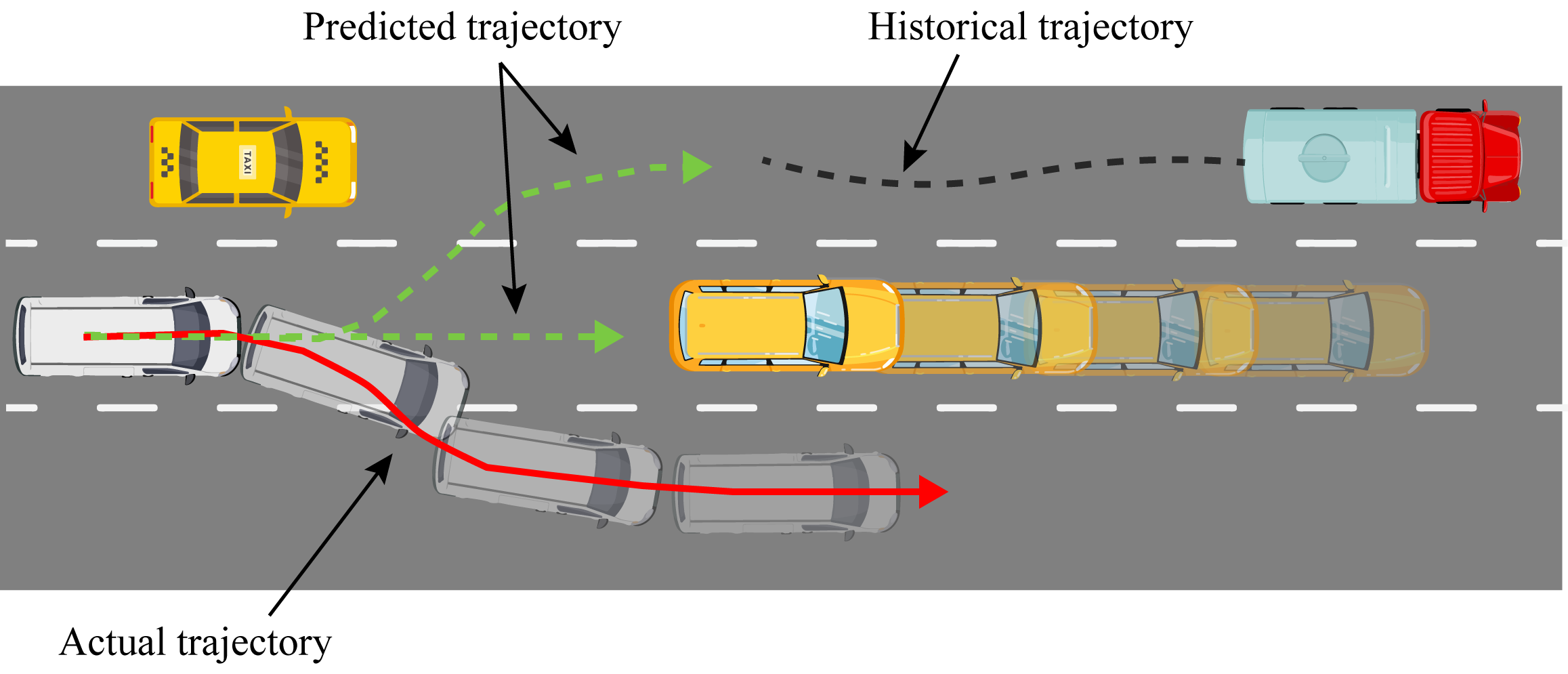}
\caption{Trajectory Prediction Example}
\label{fig:trajectory_example}
\end{figure}

Although the autonomous driving car is fully committed to liberating human from the boring driving activities, its safety and the efficiency are still the primary concerns. 
The US National Highway Traffic Safety Administration (NHTSA) breakdowns the autonomous driving cars into six categories \cite{blain2017self}. To achieve the full automation, which is defined as level 5, all the autonomous driving companies concentrate on developing their own autonomous driving system (ADS) or advanced driver-assistance systems (ADAS). However, even the best ADS only reach the conditional automation, which is level 3 in the six categories.
The trajectory prediction plays a pivotal role in any level of the autonomous driving system (ADS). In real road traffic, the number of different possible trajectories a car can take in just a few seconds may be countless. 
Figure ~\ref{fig:trajectory_example} illustrates some examples. 
A precise trajectory prediction helps the autonomous car take the correct action in next stage. 

Trajectory prediction is a challenging problem, because it does not only depend on the historical information of the target vehicle, but also the historical information of the surrounding vehicles. 
In recent years, lots intricate problems become solvable as the growth of the deep learning in many fields, such as computer vision \cite{ylu01} \cite{ylu04} \cite{ylu05}, natural language processing \cite{zhao2018learning}, intelligence hardware \cite{yuan2017memristor} \cite{fang2019event}, etc. Many papers have been published to improving the trajectory prediction algorithm \cite{houenou2013vehicle} \cite{jin2020simulation} \cite{zhao2019temporal}. 
Despite the efforts, the accuracy of the existing prediction models is not high enough. This is because the surrounding vehicles will also respond to its environment and adjust its trajectory accordingly. Without considering this, the model can not make an accurate trajectory prediction. Hence it is necessary to have information from all neighboring vehicles, and consider the potential evolvements of their trajectories in the near future. For example, the human drivers observe and surmise other drivers’ latent intention from the mirrors of the car. To emulate this behavior, an information sharing network should be established among all vehicles. 

In the rest of the paper, we use the name “\textit{target vehicle}” to refer to the vehicle whose trajectory is to be predicted and use the name “\textit{neighbors}” to refer to the surrounding vehicles. We propose a novel graph-based information sharing network (GISNet), which allows the target and neighbor vehicles to propagate and learn the trajectory features among themselves. Our proposed network is evaluated using the NGSIM highway vehicle trajectory dataset. The RMSE of the prediction is compared to several existing models. Compared with other existing trajectory prediction methods, our approach can reduce the prediction error by up to 50.00\%.
The following summarizes the major contributions of our work:

\begin{itemize}
\item A new trajectory prediction model is developed which grants the information sharing among the graph neural network.

\item The prediction is based on the embedding feature, which is derived from multi-dimensional input sequences including the historical trajectory of target and neighboring vehicles, and their relative social positions.

\item The model allows us to consider the latent intention of surrounding neighbors during the prediction. Compared with other existing trajectory prediction methods, our approach can reduce the prediction error by up to 50.00\% and achieve the state-of-the-art performance.
\end{itemize}

The rest of the paper is structured as follows: In Section \ref{sec:related}, we review the existing methods, from which we got the inspirations. This is followed in Section \ref{sec:method} by details about our GISNet. Section \ref{sec:experiments} and Section \ref{sec:results} describe our experimental steps and evaluation of the results. Finally, Section \ref{sec:conclusion} concludes this work and discusses our future works.

\section{Related Works} \label{sec:related}

Over the past several years, the autonomous driving car have played an increasingly critical role in many areas \cite{huang2018apolloscape} \cite{otaki2019autonomous} \cite{wang2018networking}. Lots of researchers dedicate their attention to accurate vehicle trajectory prediction \cite{lefevre2014survey} \cite{zhao2019simulation}. Due to space limitation, in this section, we focus on the pros and cons of more recent works in this area.

% The Long Short-term Memory (LSTM) network is design for learning the long-term dependencies.
% This network was suitable for memorizing the trajectory of vehicle. 

The Long Short-term Memory (LSTM) network is an effective model to memorize the trajectory of vehicles. Several works took the LSTM as the network backbone for the trajectory prediction \cite{shi2018lstm} \cite{altche2017lstm} \cite{xue2018ss}\cite{kim2017probabilistic}. In \cite{nikhil2018convolutional}, the authors proposed a convolutional neural network (CNN) to replace the LSTM. Their work aimed at increasing the model parallelism and efficiency. It shows that, without the LSTM operation, the efficiency of the model can be improved significantly. However, the input of the model is still formulated as the sequence-to-sequence format. The historical trajectory input are embedded into a fixed size through the fully connected (FC) layers. In order to preserve the temporal information, the historical trajectory data are stacked together follow by their time sequence order. 

Recently, many researchers started to investigate the relationship between the target vehicle and its surrounding neighbors \cite{deo2018would}\cite{li2019grip}. Hand-crafted features were integrated into the model for trajectory prediction \cite{bahram2016combined}\cite{choi2013understanding}\cite{choi2012unified}. Nonetheless, the performance of the motion prediction is highly depend on the quality of the hand-crafted features. The method of social pooling was first proposed in \cite{lefevre2014survey}. The interactions among all individuals can be shared between multiple LSTMs through the social pooling layer. As an extension of the original social pooling, the convolution operation was introduced into the model in \cite{deo2018convolutional}. 
The LSTM layer encode the historical trajectory of each vehicle into a feature vector. Each encoded trajectory feature was placed into the corresponding location in a 3D tensor which is the same as its location in the background scene. Finally, all the features were constructed as a 3D tensor. Therefore, the reception field in the convolution operation can explore the interaction between each objects. 
The non-local multi-head attention mechanism was invented to combine the relevant neighbor information \cite{messaoud2019non}. 
% The \begin{math} N_head \end{math} interaction vector are introduced. 
The model divides the road environment into grids. The learned attention weight specifies the amount of attentions that need be placed on the trajectory features associated to specific grids during the prediction. Instead of considering the interactions among neighboring objects, \cite{zhao2019multi} considers the relationship between the object and its scene background. The authors concatenated the multi-agent encoding and the scene context encoding as the input of the trajectory prediction network. The predicted trajectory was regulated by the constraints which was learnt from the scene background. 
% \cite{li2019grip} had the similar idea with us that using the graph-based network. But, the author used the normalized coordinates instead of the embedded features to represent each vehicle, this approach might reduce the feature representation ability. Furthermore, the time information was discretized in the model. Alternately, the author assumed that the links between each graph keeps the temporal information.  

In this work, we focus on enhancing the information sharing between vehicles. By adopting Graph Convolutional Network (GCN), the vehicles can learn the latent intention of its surrounding neighbors. The experimental results show that our proposed approach achieves the state-of-the-art performance.

\section{Methods}\label{sec:method}

Our proposed network is an end-to-end model that each module is fully differentiable. The loss is calculated by measuring the difference between the predicted trajectories and the ground truth trajectories. In this section, each component of the network will be elaborated.

\begin{figure*}[htb]
\centering
\includegraphics[width=\textwidth]{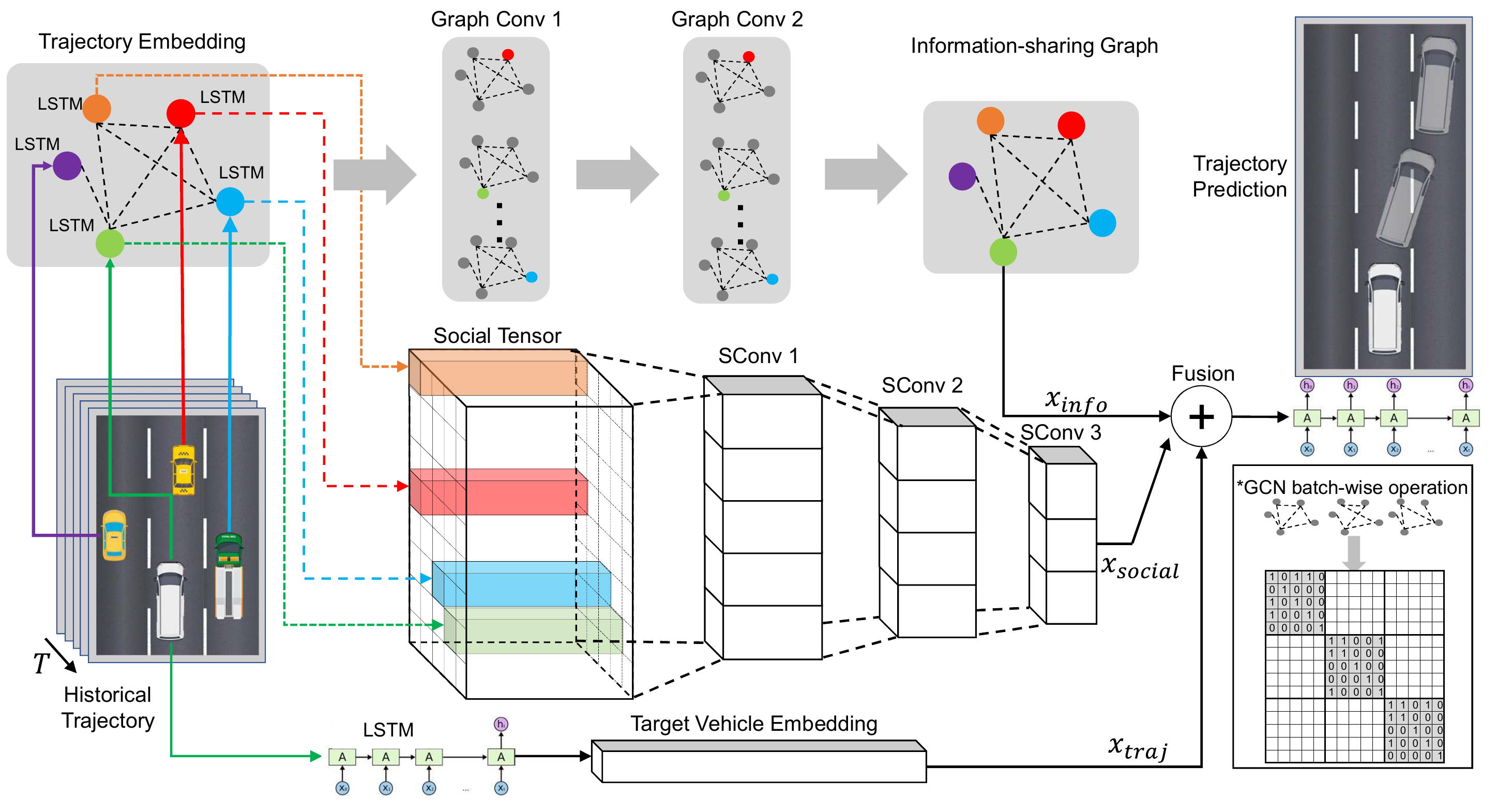}
\caption{Graph-based Information Sharing Network (GISNet) Architecture}
\label{fig:architecture}
\end{figure*}

\subsection{Historical Trajectory Formulation}

% The whole scene is divided into \begin{math} m * n \end{math} grids. The prediction-centric vehicle is located at the center of the whole map. And the width of each grid is 15 feet. In this paper, the \begin{math} m \end{math} and \begin{math} n \end{math} are defined as 13 and 3 which are the same as \cite{deo2018convolutional}. Therefore, t
The historical trajectory of each vehicle is formulated as a sequence:
\begin{equation}
     X_{coor} = \{C^{t - 1}, C^{t - 2}, ... , C^{t}\}  
\end{equation}
where, 
\begin{equation}
     C^{t} = {x^t,y^t}  
\end{equation}
is the collection of the historical trajectory coordinates which contain \begin{math} x \end{math}, \begin{math} y \end{math} values. The \begin{math} t \end{math} is the time horizon of the historical trajectory which is set to be 3 second in this paper.

\subsection{Vehicle Information Embedding}

The LSTM model has the ability to memorizing the long term dependency from the past information. So, it can be used for extracting the features from the vehicle's historical trajectories, as shown in the left part of Figure~\ref{fig:architecture}. The LSTMs which are used for extracting the embedding trajectory features from all vehicles share the same weight. In this way, the hidden states of all vehicles have consistent representations. For each vehicle \begin{math} i \end{math}, a 1-d embedding vector \begin{math} x_{traj} \end{math} with size  \begin{math} l \end{math} is extracted by the LSTM, which captures the trajectory features of the vehicle. It will be placed into a \begin{math} m * n \end{math} grid system to from a 3D tensor. The placement position is determined by the grid location of the vehicle in the scene background. In this paper, the \begin{math} m \end{math} and \begin{math} n \end{math} are defined as 13 and 3 which are the same as \cite{deo2018convolutional}. Hence, the relative positional relationship of each vehicle can be preserved. Those grids that do not have a vehicle will be filled with zeros. The tensor is processed by a convolutional layers followed by pooling layers to extract the 1-d feature vector, \begin{math} x_{social} \end{math}. It contains the social relationships among the vehicles, as shown in the middle part of Figure~\ref{fig:architecture}.

At the same time, we keep a separate copy of the trajectory feature of the target vehicle, \begin{math} x_{traj} \end{math}, as shown at the bottom of Figure~\ref{fig:architecture}. It will be integrated with the \begin{math} x_{social} \end{math} later in the prediction stage.

\subsection{Graph-based Information Sharing}
After the LSTM encoding, the embedding collection of all vehicles' historical trajectory is generated, which is denoted as:
\begin{equation}
     X_{vehicles} = \{x_{traj}^1, x_{traj}^2, ... , x_{traj}^n\}  
\end{equation}
where \begin{math} n \end{math} is the number of vehicles that can be observed in the current scene. For better prediction, the target vehicle needs to learn the latent intention from its surrounding neighbors. However, the structure of the information sharing network are generated from the non-euclidean domain. Compared with the traditional CNN model, the graph convolutional network has its talent in exploring the meaningful features from the irregular structures \cite{kipf2016semi}.

The equation of information propagation between layers in the GCN is defined as following:
\begin{equation}
     H^{(l+1)} = \sigma(\hat{D}^{-\frac{1}{2}}\hat{A}\hat{D}^{-\frac{1}{2}}H^{(l)}W^{(l)})  
\end{equation}
where,
\begin{equation}
     \hat{A} = A + I  
\end{equation}
\begin{math} \hat{A} \end{math} is the matrix contains the adjacent matrix \begin{math} A \end{math} and the identity matrix \begin{math} I \end{math}. The \begin{math} \hat{A} \end{math} matrix has two objectives: 1. allow the information sharing between the node and its adjacent neighbors. 2. each node can consider the lower level feature from itself. The \begin{math} \hat{D} \end{math} is the matrix describes the degree of each node. The feature of each node at layer \begin{math} l \end{math} is defined as \begin{math} H^{(l)} \end{math}. For the input layer, the \begin{math} H \end{math} is equal to \begin{math} X \end{math}. The \begin{math} \sigma \end{math} is the activation function to improve the representability of the model.

In this paper, we apply a two-layer graph convolutional network (GCN). The feature size in both layers is set to 64. The \begin{math} \hat{A} \end{math} is a zero-one matrix. Its \begin{math} ij \end{math}th entry (\begin{math} \hat{a_{ij}} \end{math}) is zero if there is no connection between vertices \begin{math} i \end{math} and \begin{math} j \end{math}, otherwise it is 1. In our case, the vehicle which is being predicted is connected with all surrounding vehicles. 
Therefore, the forward path of the model is defined as following:
\begin{equation}
     f(X, A) = (\hat{D}^{-\frac{1}{2}}\hat{A}\hat{D}^{-\frac{1}{2}}ReLU(\hat{D}^{-\frac{1}{2}}\hat{A}\hat{D}^{-\frac{1}{2}}XW^{(0)})W^{(1)})
\end{equation}
where the \begin{math} ReLU \end{math} is the activation function between layers. The \begin{math} W^{(0)} \end{math} and \begin{math} W^{(1)} \end{math} are the parameters within the two graph convolution layers. 
The architecture of the information sharing module is given in the top part of Figure~\ref{fig:architecture}. After the two layer convolution operations, a 1-d feature \begin{math} x_{info} \end{math} that summarizes the latent intention of surrounding neighbors is acquired. 

In order to improve the training efficiency, we employ the batch-wise operation for multi graph in the GCN, as shown in the bottom right of Figure~\ref{fig:architecture}. First, all graphs are concatenated together to build a fusion graph. Then, a fusion block diagonal matrix is established. Each one of them represents the connectivity of the graph instance.

\subsection{Future Trajectory Generation}

Finally, all three features, \begin{math} \{x_{traj}, x_{social}, x_{info}\} \end{math}, are concatenated together to construct an embedding of the vehicle future trajectory, which is to be predicted. Then Then, the embedding feature is passed to the trajectory generation module, as shown in the top right of Figure~\ref{fig:architecture}. In the generation module, the LSTM based decoder is applied to generate the sequence of \begin{math} x, y \end{math} coordinates for the next 5 cycles, which is the prediction horizon that we are interested. The output of the model is denoted as a sequence: 
\begin{equation}
     Y_{coor} = \{C^{t - 1}, C^{t - 2}, ... , C^{t}\}  
\end{equation}
where, 
\begin{equation}
     C^{t} = {x^t,y^t}  
\end{equation}
is the predicted \begin{math} x \end{math}, \begin{math} y \end{math} values of the target vehicle at time \begin{math} t \end{math}.

\section{Experiments}\label{sec:experiments}

\subsection{Dataset}\label{sec:data}

\begin{figure}[!htbp]
\begin{subfigure}{.25\textwidth}
  \centering
  \includegraphics[width=1.7in]{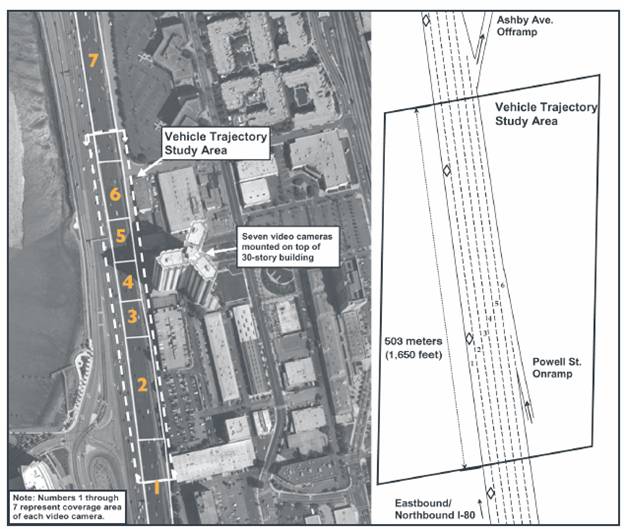}
  \caption{I-80 study area}
  \label{fig:i80}
\end{subfigure}%
\begin{subfigure}{.25\textwidth}
  \centering
  \includegraphics[width=1.7in]{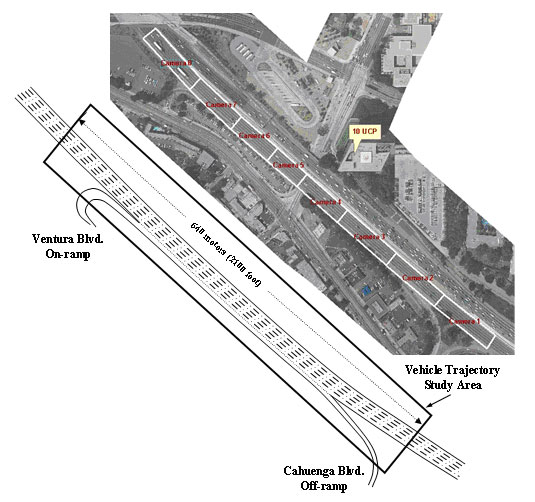}
  \caption{I-101 study area}
  \label{fig:i101}
\end{subfigure}
\caption{Data Collection Procedure}
\label{fig:data}
\end{figure}

Our proposed model is evaluated on the public Next Generation Simulation (NGSIM) dataset. 
The NGSIM dataset collects the detailed vehicle trajectory information on eastbound I-80 in the San Francisco Bay area \cite{colyar2006us} and southbound US 101 in Los Angeles \cite{colyar2007us}. 
The study area for I-80 and 101 are 500 meters (1,640 feet) and 640 meters (2,100 feet), respectively. Figure \ref{fig:data} shows the study area of the NGSIM dataset. All the data is segmented into three 15 minute periods. 
The dataset is splitted into three subsets: training, validation and testing. We follow the same approach in \cite{deo2018convolutional} to split the vehicle trajectories into 8 second segments. The first 3 seconds are treated as the historical data, and the following 5 seconds trajectories are to be predicted.

\subsection{Evaluation Metrics}

In this paper, we use the Root Mean Square Error (RMSE)  \cite{chai2014root} to evaluate the performance of our proposed approach. This metric measures the difference between the predicted trajectory and the observed trajectory (ground truth) using the following equations: 
\begin{equation}
     RMSE = \sqrt{\frac{1}{n}\Sigma_{m=1}^{n}{{(x_m^T - x_m^{\prime}{}^T)}{}^2 + {(y_m^T - y_m^{\prime}{}^T)}{}^2}}
\end{equation}
where \begin{math} x_m^t \end{math} and \begin{math} y_m^t \end{math} are the predicted coordinates. The \begin{math} m \end{math} is the index of sample. The total number of testing sample is denoted as n, and  \begin{math} x_m^{\prime}{}^t \end{math} and \begin{math} y_m^{\prime}{}^t \end{math} are the ground truth coordinates. \begin{math} T \end{math} is the prediction horizon of the model. In our experiment, \begin{math} T \end{math} is varying from 1 to 5 seconds.

\subsection{Comparison Baselines}

\begin{itemize}
\item Constant Velocity (CV) \cite{schneider2013pedestrian}: A baseline method that uses the constant velocity (CV) Kalman filter to forecast vehicle trajectory .

\item GAIL-GRU \cite{kuefler2017imitating}: A generative adversarial imitation learning model that takes the ground truth trajectories of all adjacent neighbors as the model input.

\item Vanilla LSTM (V-LSTM): The typical LSTM based encoder-decoder model. The vehicle historical trajectory is fed into the model as the input. Then, the LSTM output is decoded as the vehicle trajectory prediction.

\item Social-LSTM (S-LSTM) \cite{alahi2016social}: The model applies the social pooling layer which allows the information sharing between each individual LSTM.

\item ConvSocial-LSTM (CS-LSTM) \cite{deo2018convolutional}: The model uses the convolution operation to extract the features from the social tensor. The prediction-centric vehicle’s feature is concatenated with social feature.

\item Non-local Social Pooling (NLS-LSTM) \cite{messaoud2019non}: This model is based on an LSTM encoder-decoder. The social pooling is applied to capture the interactions between all vehicles. Besides, non-local multi-head attention mechanism is used to summarize the relevant information.

\item Multi-Agent Tensor Fusion (MATS) \cite{zhao2019multi}: This model concatenates the background scene feature and the vehicle historical trajectory feature into a multiagent tensor. A generative adversarial networks (GAN) based module is included for generating the future trajectory prediction.

\end{itemize}

\subsection{Experiment Setup}\label{sec:experiment_setup}
We run our experiments on a desktop server running Ubuntu 16.04 OS with 3.60GHz Intel Xeon W-2123 CPU, 256GB Memory and a NVIDIA 2080Ti GPU. During the training, the Adam optimizer is applied with a 0.001 learning rate. The graph-based information sharing model has a 64 dimensional embedding state. We use the ReLU to be the activation function. Batch normalization and dropout are also applied for preventing the overfitting. The training and testing framework is built in PyTorch.

\section{Results}\label{sec:results}

%%%%%%%%%%%%%%%%%%%%%%%%%%%%%%%%%%%%%%%%%%%%%%%%%%%%%%%%%%%%%%%%%%%%%%%%%%%%%%%%%%%%
% table and figure for the result section
\begin{table*}
\begin{center}
{\caption{The evaluation of RMSE in meters on NGSIM dataset}\label{table1}}
\begin{tabular}{lccccccccc}
\hline
\rule{0pt}{12pt}
Horizon (s) & CV & GAIL-GRU & V-LSTM & S-LSTM & CS-LSTM & NLS-LSTM & MATS & GISNet
\\
\hline
\\[-6pt]
\quad 1 & 0.73 & 0.69 & 0.68 & 0.65 & 0.61 & 0.56 & 0.66 & \textbf{0.33}\\
\quad 2 & 1.78 & 1.51 & 1.65 & 1.31 & 1.27 & 1.22 & 1.34 & \textbf{0.83}\\
\quad 3 & 3.13 & 2.55 & 2.91 & 2.16 & 2.09 & 2.02 & 2.08 & \textbf{1.42}\\
\quad 4 & 4.78 & 3.65 & 4.46 & 3.25 & 3.10 & 3.03 & 2.97 & \textbf{2.14}\\
\quad 5 & 6.68 & 4.71 & 6.27 & 4.55 & 4.37 & 4.30 & 4.13 & \textbf{3.23}
\\
\hline
\\[-6pt]
\end{tabular}
\end{center}
\end{table*}

\begin{figure*}[!htbp]
\centering
\begin{subfigure}{0.45\linewidth}
  \centering
  \includegraphics[width=\linewidth]{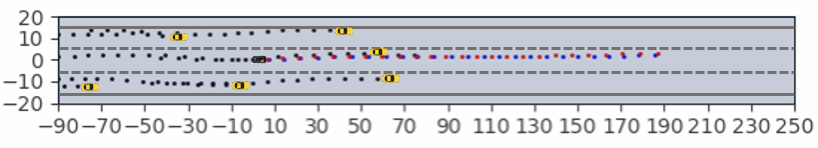}
  \caption{Fast Speed Traffic}
  \label{fig:traj_1}
\end{subfigure}
\begin{subfigure}{0.45\linewidth}
  \centering
  \includegraphics[width=\linewidth]{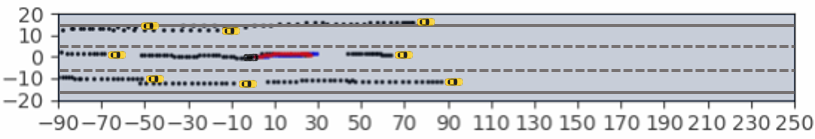}
  \caption{Low Speed Traffic}
  \label{fig:traj_2}
\end{subfigure}

\begin{subfigure}{0.45\linewidth}
  \centering
  \includegraphics[width=\linewidth]{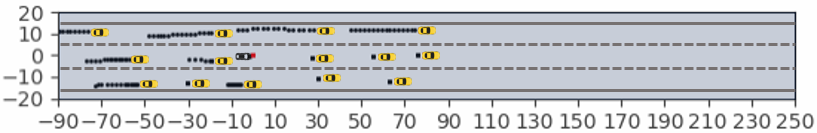}
  \caption{Congested Traffic}
  \label{fig:traj_3}
\end{subfigure}
\begin{subfigure}{0.45\linewidth}
  \centering
  \includegraphics[width=\linewidth]{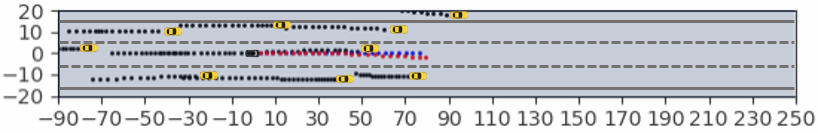}
  \caption{Crowded Traffic}
  \label{fig:traj_4}
\end{subfigure}
\caption{Trajectory Prediction Visualization}
\label{fig:traj}
\end{figure*}

%%%%%%%%%%%%%%%%%%%%%%%%%%%%%%%%%%%%%%%%%%%%%%%%%%%%%%%%%%%%%%%%%%%%%%%%%%%%%%%%%%%%

\subsection{Predicted Trajectory Accuracy Improvement}

Table \ref{table1} shows the RMSE results for the models being compared. The first thing we can notice is that all the deep learning based methods outperform the traditional model (CV). It demonstrates the efficiency of the deep learning based model. The vanilla LSTM considers the temporal trajectory of the target vehicle. And the generative adversarial imitation learning - gated recurrent unit (GAIL-GRU) extends the lstm architecture by importing the GAN. However, none of them consider the impact of the neighbor cars. Hence, they also perform poorly in the prediction.

The second thing we can observe is that, all the models which consider the surrounding vehicles give the lower RMSE. It proves that the neighbors information does help the vehicle trajectory prediction. Moreover, our proposed method outperforms all other baselines due to the information-sharing mechanism. Compared with the original convolutional social pooling (CS-LSTM) method, we can achieve 45.35\%, 34.02\%, 31.69\%, 30.74\% and 25.86\% accuracy improvements when prediction horizon varies from 1s to 5s. Compared with the MATS, our model also achieves 50.00\%, 38.06\%, 31.73\%, 27.95\% and 21.79\% accuracy improvements for the 5 prediction horizons. Although CS-LSTM and MATS both considers the features from neighboring vehicles, they were placed in the social tensor and processed by a convolutional neural network. The GISNet outperforms MATS and CS-LSTM due to two reasons: 1) Small CNN kernels are used in these model, therefore, they only consider the joint features of vehicles in adjacent area. Although the covered area of the joint features increases as the network goes deeper, the resolution of the information is also reduced due to the pooling layers. 2) The social tensor does not only have the useful features of the neighboring vehicles, but also has lots of empty features located at the grid location not occupied by any vehicles. 
Finally, compared with the NLS-LSTM, the proposed model reduces the RMSE by 41.07\%, 31.97\%, 29.70\%, 29.37\% and 24.88\% in different prediction horizons, respectively. One reason for this is that the non-local multi-head attention mechanism will only extract features from the ``relative'' important surrounding vehicles. Some minor but meaningful features might be suppressed.

\subsection{Vehicle Predicted Trajectory Visualization}

In this section, we visualize several predicted trajectories and the ground truth to give a qualitative demonstration of the prediction performance. All the results are sampled from the NGSIM data set. And the data in the NGSIM is collected from the real world. We select 4 different scenarios to reproduce some typical scenes in daily-life: a) Fast speed traffic, b) Low speed traffic, c) Congested traffic, d) Crowded Traffic. The results are given in Figure~\ref{fig:traj}. In the figure, the black vehicle is the car which is being predicted, and the black vehicles are the surrounding neighbors. The black dash lines are the historical trajectory of each vehicle. The blue dash line is the trajectory predicted by model, and the red dash line is the ground truth trajectory.

As we can see that, our predicted trajectories are close to the ground truth. In the high speed traffic scenario, the cars are driving at a relatively high speed. The final location of the predicted trajectory is almost the same as the observed location, as shown in Figure~\ref{fig:traj}.a. In the low speed traffic scenario (Figure~\ref{fig:traj}.b), the GISNet learns that the vehicle is in a relative low speed, and the predicted trajectory is shorter. For the congested traffic scenario, the result is given in Figure~\ref{fig:traj}.c. We can see that all vehicles on the left most lane are moving, however, the car which is being predicted is in a congested lane. Consequently, the model is not affect by the surrounding cars and can predict the stationary trajectory. Last, the most complex situation in daily life is the crowded traffic, where the target vehicle is moving but crowded with many cars. Figure~\ref{fig:traj}.d shows the predicted result in this scenario, where the vehicle is trying to make a lane change. In this scenario, our model can still predict the motion. In general, our model can output an accurate car location for the near future (1s, 2s and 3s) and make good prediction of the trend for the long term (4s and 5s).

% \section{DISCUSSION AND FUTURE WORKS}\label{sec:conclusion} 

\section{CONCLUSION}\label{sec:conclusion} 
In this paper, we propose a novel graph-based information sharing network (GISNet). The network has the ability to encode the historical trajectory of each vehicle and allow the information sharing between the target vehicle and its surrounding neighbors. Furthermore, the model can fuse the features extracted from both Euclidean domain and non-Eculidan domain to make the future trajectory prediction. We apply our network to a public dataset to demonstrate its capability to predict an accurate trajectory in the future. Meanwhile, our method outperforms other reported trajectory prediction methods and can reduce the prediction error by up to 50.00\%. The qualitive results also demonstrate that the GISNet can capture the vehicle motion trend and generate the accurate prediction result.

In future work, we aim at exploring a paradigm to build the information sharing network for multi-vehicle trajectory prediction. The information sharing network for multi-vehicle can not be only depend on the absolute location of each vehicles at the last time stamp. The relative location and relationship of each vehicle are changing all the time. Therefore, the time information and the historical trajectory of each vehicle should be considered when building the topology communication network. 

% Furthermore, the background scene feature should also be taken into account for the trajectory prediction. In \cite{zhao2019multi}, the additional features are extracted from the image to help making a more accurate prediction. However, the general feature can not help the model, in contrast, it may introduce extra noises. Hence, a more detailed and targeted feature need to be included in model prediction, for example the road segmentation information. These features could be used to further improve the accuracy of the vehicle trajectory prediction.

\bibliographystyle{IEEEtran}
\bibliography{reference}

\end{document}